\documentclass[sigconf]{acmart}
\usepackage{listings} 
\usepackage{xurl}

\AtBeginDocument{%
  }

\setcopyright{rightsretained}
\copyrightyear{2024}
\acmYear{2024}

\begin{document}

\title{LLM-Powered Ensemble Learning for Paper Source Tracing: \\ A GPU-Free Approach}

\author{Kunlong Chen}
\email{chenkunlong.ckl@antgroup.com}

\author{Junjun Wang}
\email{xingruo.wjj@antgroup.com}

\author{Zhaoqun Chen}
\email{zhaoqun.czq@antgroup.com}

\affiliation{%
  \institution{Ant Group}
  \city{Hangzhou}
  \state{Zhejiang}
  \country{China}
}

\author{Kunjin Chen}
\email{kunjin.ckj@taobao.com}
\affiliation{%
  \institution{Alibaba Group}
  \city{Hangzhou}
  \state{Zhejiang}
  \country{China}}

\author{Yitian Chen}
\email{yitianartsky@gmail.com}
\affiliation{%
  \institution{Cardinal Operations}
  \city{Beijing}
  \country{China}}

\renewcommand{\shortauthors}{Chen et al.}

\begin{abstract}
We participated in the KDD CUP 2024 paper source tracing competition and achieved the 3rd place. This competition tasked participants with identifying the reference sources (i.e., \textit{ref-sources}, as referred to by the organizers of the competition) of given academic papers. Unlike most teams that addressed this challenge by fine-tuning pre-trained neural language models such as BERT or ChatGLM, our primary approach utilized closed-source large language models (LLMs). With recent advancements in LLM technology, closed-source LLMs have demonstrated the capability to tackle complex reasoning tasks in zero-shot or few-shot scenarios. Consequently, in the absence of GPUs, we employed closed-source LLMs to directly generate predicted reference sources from the provided papers. We further refined these predictions through ensemble learning. Notably, our method was the only one among the award-winning approaches that did not require the use of GPUs for model training. Code available at \url{https://github.com/Cklwanfifa/KDDCUP2024-PST}.
\end{abstract}

\keywords{Paper source tracing, Large Language Model}


\maketitle

\section{Introduction}
In the rapidly evolving landscape of scientific research, tracing the origins of ideas of academic papers has become increasingly crucial. The ability to accurately identify the source references of a paper not only enhances our understanding of the pathways of scientific progress but also holds significant scientific and societal value \cite{zhang2024pst}. As the volume of scientific publications continues to grow exponentially \cite{roadmap}, there is an urgent need for efficient algorithms that can swiftly identify the sources of papers. Such tools are particularly valuable for emerging researchers, enabling them to quickly grasp the developmental trajectory of specific technologies and situate their work within the broader scientific context.

Traditionally, researchers have used methods such as random forests \cite{valenzuela2015identifying} or a neural network training \cite{Yin2021} to identify key citations for each paper. However, these approaches often require substantial time and effort in designing intricate features or fine-tuning pre-trained models to achieve satisfactory performance. The advent of large language model (LLM) technologies has opened new avenues for addressing this challenge. Recent studies have demonstrated that when the parameter count is sufficiently high, LLMs can solve complex problems through zero-shot learning \cite{Kojima2022}.

In this paper, we propose a novel approach that leverages pre-trained LLMs as unsupervised reasoners to identify a paper's source references. Our method utilizes the paper's text and related information to perform direct reasoning, eliminating the need for extensive feature engineering or model fine-tuning. This approach is particularly beneficial for researchers with limited GPU resources, as it offers a balance between computational efficiency and performance accuracy.

The efficacy of our proposed method is validated by its success in the paper source tracing competition of KDD CUP 2024, where it secured third place. Notably, our team was the only award-winning entry that did not rely on GPUs, underscoring the method's ability to achieve competitive results with minimal computational resources. This achievement demonstrates that our approach effectively balances resource consumption and performance, making it a valuable tool for a wide range of researchers and institutions.

\section{PRELIMINARIES}

\subsection{Data Description}
The dataset we used in this study is from the Open Academic Graph dataset \cite{zhang2024oag}, which includes training and test sets for multiple tasks under the academic paper mining theme, such as Author Name Disambiguation, Academic question answering, and Paper source tracing. The method we employ is applicable to the data from the Paper source tracing task in the OAG dataset \cite{zhang2024pst}, which is referred to as the PST dataset in the following text. The PST dataset contains 1576 papers related to computer science and 55014 associated citations.


\subsection{Task Description}
The primary objective of our study is to identify the most significant reference paper for a given academic work. To formalize this task, we introduce the following notation.

\begin{itemize}
    \item Let $\mathcal{A}$ denote the set of all references that have been cited at least once across the entire dataset.
    \item For each paper $P_i$, we define $R_i = [r_{i,1}, r_{i,2}, \cdots, r_{i, J_i}]$ as the set of its references, where each $r_{i,j} \in \mathcal{A}$.
    \item We denote $S_i \subset R_i$ as the subset of references that serve as source references for paper $P_i$.
\end{itemize}

Our task is to assign a probability score $p_{i,j} \in [0, 1]$ to each reference $r_{i,j}$ in $R_i$, where higher values indicate a greater likelihood of the reference being a key source for the paper. To evaluate the performance of our model, we employ the following metrics:

\begin{itemize}
    \item For each paper $P_i$, we calculate the Average Precision (AP), denoted as $\text{AP}_i$.
    \item The overall model performance is assessed using the Mean Average Precision (MAP), computed as $\text{MAP} = \frac{1}{N} \sum_{i=1}^N{\text{AP}_i}$, where $N$ is the total number of papers in the dataset.
\end{itemize}

\section{METHOD}

Given the constraints of limited access to high-performance GPUs, we developed an innovative approach that diverges from traditional methods such as fine-tuning large pre-trained models or employing graph machine learning techniques. Our methodology primarily leverages closed-source LLMs in conjunction with feature engineering and ensemble learning to address the paper source tracing problem effectively.

\subsection{Feature Engineering}

While our method relies heavily on LLMs for text comprehension and problem-solving, we found that incorporating carefully engineered features from the papers enhances performance. Our feature extraction process focuses on the following key aspects:

\begin{enumerate}
    \item Paper Metadata: We extract information such as the publication venue (journal/conference name), citation count, and author details (nationality and affiliation) for each paper.
    
    \item Citation Statistics: We compute the total occurrences of each cited reference within the paper, as well as its distribution across different sections or chapters.
    
    \item Reference Metadata: Similar to the main paper, we extract publication venue and author information for each cited reference.
    
    \item Contextual Keywords: We count the occurrences of specific keywords (e.g., \textit{motivated by}, \textit{inspired by}) in the vicinity of each citation to gauge its importance.
\end{enumerate}

These features provide a rich, structured representation of the papers and their citations, complementing the semantic understanding capabilities of the LLMs.

\subsection{LLM-generated Answers}

To generate diverse and high-quality answers, we employ four state-of-the-art LLMs: GPT-4 Turbo, GPT-4o, Gemini 1.5 Pro, and Claude 3 Opus. Our selection criterion was based on empirical evidence suggesting that LLMs with an MMLU (Massive Multitask Language Understanding) score above 85 demonstrate superior logical reasoning capabilities in academic contexts.

\subsubsection{Prompt Engineering}

We designed a set of prompts to elicit nuanced responses from the LLMs:

\begin{enumerate}
    \item Base prompt: A carefully crafted instruction set (illustrated in Figure 1) that guides the LLM in identifying source citations.
    
    \item Inspiration-focused prompt: This variant asks the LLM to categorize citations as "direct inspiration", "indirect inspiration", or "other inspiration", emphasizing the keyword "inspiration".
    
    \item Title-enriched prompt: An extension of the base prompt that includes the titles of cited articles for additional context.
    
    \item Meta-optimized prompt: We utilize GPT-4 to refine the base prompt itself, leveraging the LLM's meta-learning capabilities.
    
    \item Notes-based prompt: For papers with available "notes" fields (potentially containing annotators' insights), we instruct the LLM to identify key citations based on these descriptions.
\end{enumerate}

\subsubsection{Answer Generation}

Let $N_{LLM}$ denote the number of LLM types and $N_{prompt}$ the number of prompt variants. Theoretically, for each academic paper, we could generate up to $M_p = N_{LLM} \cdot N_{prompt} \cdot M$ answers, where $M$ is the maximum number of answers per LLM-prompt combination. However, due to practical constraints, we limit this to $M_p$ answers per paper.

For each cited reference $r_{i,j}$ in the citation set $R_i$ of paper $P_i$, we generate a probability list $P_{i,j}^{LLM}$ of length $M_p$, where each element represents the LLM-estimated probability of $r_{i,j}$ being a source citation. Formally, we have:
\begin{equation}
    P_{i,j}^{LLM} = [p_1, p_2, \ldots, p_{M_p}], \quad p_k \in [0, 1]
\end{equation}

\subsection{Base Models}

To complement the LLM-generated probabilities, we employ two gradient boosting frameworks: LightGBM \cite{ke2017lightgbm} and CatBoost \cite{prokhorenkova2018catboost}. These models operate on the engineered features described in Section 3.1.

For each paper-citation pair $(P_i, r_{i,j})$ in the training set, we construct a feature vector $X_{i,j}$ and a corresponding binary label $Y_{i,j} \in \{0, 1\}$, where $Y_{i,j} = 1$ indicates that $r_{i,j}$ is a reference source of $P_i$. We train two classifiers:
\begin{equation}
    f_{lgb}(X_{i,j}) \rightarrow [0, 1] \quad \text{(LightGBM)}
\end{equation}
\begin{equation}
    f_{cb}(X_{i,j}) \rightarrow [0, 1] \quad \text{(CatBoost)}
\end{equation}
These classifiers provide probability estimates for each citation being a source citation based on the engineered features.

\subsection{Ensemble Approach}

Our final prediction model combines the strengths of LLM-generated answers and traditional machine learning classifiers. For each paper-citation pair $(P_i, r_{i,j})$, our goal is to find a function that combines the prediction results of different methods as:

\begin{equation}
    P_{i,j} = f_{ensemble}((P_{i,j}^{LLM}) , f_{lgb}(X_{i,j}) , f_{cb}(X_{i,j}))
\end{equation}
This ensemble approach allows us to leverage the semantic understanding of LLMs while benefiting from the structured feature representations captured by traditional models.

\subsection{Our framework}

In this section, we present our complete approach. The algorithm takes as input the main text of paper $P_i$ and its list of cited references $R_i$, and outputs the probability $p_{i,j}$ for each reference $r_{i,j}$ being a source citation of paper $P_i$.

\begin{enumerate}
\item Using $f_{lgb}$ and $f_{cb}$ trained on the training set, we score each reference $r_{i,j}$, where the score represents the probability of it being a source citation. This score is represented as $p_{i,j}^{\text{base}} = p_{i,j}^{lgb} \times \omega^{lgb} + p_{i,j}^{cb} \times \omega^{cb}$.

\item We modify $p_{i,j}^{\text{base}}$ using the results from LLMs. This process involves two main steps:
    \begin{itemize}
        \item First, we group the LLM outputs into $N^g_{LLM}$ groups based on prompt type and base model type. Within each group, we ask each LLM to provide a confidence score for its predicted key citations, as represented by the prob2score function (Equation~\ref{eq:prob2score}). Additionally, we assign a weight $\omega^{llm}_{i}$ to each group based on expert knowledge, representing the importance of that LLM group. Finally, we aggregate the results of all LLM groups according to their weights to obtain $\text{score}^{\text{bonus}}_{i,j}$, which serves as a bonus to be added to $p_{i,j}^{\text{base}}$.
        \item In the second step, for each candidate reference, we aggregate the results from all LLMs. If the score given by the LLM at the $p_{neg}$-th percentile is less than $p_{threshold}^{neg}$, we reduce the bonus magnitude from the previous step. Our approach is to divide $\text{score}^{\text{bonus}}_{i,j}$ by a constant $C_{neg}$.
    \end{itemize}

\begin{equation}
    \label{eq:prob2score}
    \text{prob2score}(x) = 
    \begin{cases}
        3, & \text{if } x \geq 0.9 \\
        2, & \text{if } 0.5 \leq x < 0.9 \\
        1, & \text{if } 0.4 < x < 0.5 \\
        0, & \text{if } x \leq 0.4
    \end{cases}
\end{equation}

\item Our final prediction can be expressed as:
    \begin{equation}
        p_{i,j}^{\text{final}} = p_{i,j}^{\text{base}} + \omega_f \cdot \text{score}^{\text{bonus}}_{i,j}
    \end{equation}
    where $\omega_f$ is the weight used to adjust the base score using the LLM prediction results.
\end{enumerate}

\section{Result}
We present the performance of our proposed method on the validation set in Table~\ref{tab:score}. As can be observed, using LightGBM or LLM individually does not yield high scores. When the two approaches are combined, however, a significant improvement in the score can be achieved. This demonstrates the effectiveness of ensemble learning techniques \cite{zhou2021ensemble}. In addition, we present the configuration of parameters in Table~\ref{tab:score}.

\begin{table}
\caption{MAP scores on validation set}
  \label{tab:score}
  \begin{tabular}{lc}
        \hline
        \textbf{Method} & \textbf{Score} \\
        \hline
        SciBERT (baseline) & 0.29 \\
        Our method (only LightGBM) & 0.42  \\
        Our method (only LLM) & 0.45  \\
        Our method (combined) & 0.50  \\
        \hline
  \end{tabular}
\end{table}

\begin{table}
\caption{The configuration of parameters}
  \label{tab:params}
  \begin{tabular}{lc}
        \hline
        \textbf{Parameters} & \textbf{Value} \\
        \hline
        $\omega^{lgb}$ & 0.4 \\
        $\omega^{cb}$ & 0.6 \\
        $\omega_f$ & 0.035 \\
        $p_{threshold}^{neg}$ & 0.2\\
        $C_{neg}$ & 4 \\
        $p_{neg}$ & 0.4 \\
        \hline
  \end{tabular}
\end{table}

\section{Conclusion}
In this paper, we have briefly introduced our method that secured third place in the KDD CUP 2024 paper source tracing competition. Our approach leverages the zero-shot capabilities of LLMs, achieving impressive performance without the need for GPU-intensive model training. We demonstrate that ensemble learning techniques can significantly enhance the zero-shot performance of LLMs.

We believe that our work highlights the potential of combining multiple LLMs to harness their collective intelligence for complex reasoning tasks. This approach is particularly valuable in resource-constrained environments, as it enables sophisticated inference without the need for extensive computational resources.

Looking ahead, we encourage the machine learning community to focus on developing methods that effectively integrate the strengths of diverse LLMs for inference tasks. Such research directions hold promise for expanding the applicability of advanced AI systems across a wide range of domains and computational settings.

\begin{figure*}[t]
\centering
\begin{lstlisting}[basicstyle=\ttfamily,breaklines=true,frame=single]
def get_prompt_gpt(text):
    return f'''**** I have a task to identify the source papers of a given paper, which author indicates that they inspire them most based on its text. 
                **** I will now give you a text of academic papers, to find the most pertinent source papers:
                Firstly, Determine the primary challenges outlined in the paper, and understand the algorithm proposed by the author.
                Then, look for key phrases such as "inspired by", "motivated by", "inspired us", "motivated us", "take inspiration", "the pioneering/previous work", "following.. we adopt ... to solve the challenge/problem", "we use... based on to achieve..." or other language that indicates a strong reliance on previous research for developing the paper's core contributions.
                If key phrases exist, locate the key phrases in the text and find the sources papers which are indicated by these key phrases.
                If key phrases do not exist or for other reasons, identify the novel methods and approaches the paper introduces to tackle these challenges and locate references that are directly linked to these main challenges and methods.
                Verify that your answer do not include the ref papers appearing at the begining part of the text which describe the historical findings  like "someone et al. proposed...", normally they are not direct related to the paper's topic.
                Verify that the source papers are directly relevant to the paper's novel contributions very directly.
                Specifically highlight any references that are preceded by phrases indicating direct inspiration or motivation, such as 'Inspired by [reference]', and make these references a priority in the list
                Please provide a concise list of source papers based on the aforementioned criteria, ideally limiting the selection to the most central references that heavily influenced the main contributions of the paper. 
                **** Normally you should return less than 8 source papers. ****
                **** Please re-evaluate your result by the following metric: 		
                * Is the main idea of paper p inspired by the reference?
                * Is the core method of paper p derived from the reference?
                * Is the reference essential for paper p? Without the work of this reference, paper p cannot be completed. 
                Then, you should return your result in the json format, with the key is "reference number" and the value is "confidence score" between 0 and 1.
                **** The text of the paper is:{text}'''
\end{lstlisting}
\caption{The basic prompt for generating the answer.}
\end{figure*}

\bibliographystyle{ACM-Reference-Format}
\bibliography{sample-base}

\appendix

\end{document}